\documentclass{amsart}
\usepackage{amssymb, latexsym}

\begin{document}

\title{A New Solution to the Optimal Seeding Problem}
\author{Rita Gitik} 
\address{ Department of Mathematics \\ University of Michigan \\ Ann Arbor, MI, 48109}  
\email{ritagtk@umich.edu}

\date{\today}

\begin{abstract}
A new version of generation theory which is used to derive a new, remarkably simple, solution to the optimal seeding problem, is introduced.
This novel approach is illustrated by applying it to the problem of creating robot colonies in space.
\end{abstract}

\keywords{Algorithm, Graph, Self-Replication, Optimal Seed, Generation Theory, Robot Colony}

\subjclass[2010]{Primary: 05C85; Secondary: 68R10, 05C21}
\maketitle
\section{Introduction}

Recent research for space exploration, such as \cite{Fo}, \cite{Met}, and \cite{MMMM}  considers creating robot colonies on extraterrestrial objects for mining and manufacturing. However, sending such colonies through space is expensive \cite{W-L}, and one way to reduce the cost is to transport only one colony capable of self-replication. 

The study of self-replicating systems was initiated by John von Neumann, who discussed cellular automata capable of building other cellular automata in \cite{von New}. The foundations of the automata theory were discussed by Claude Shannon in \cite{Shannon}. The theory of self-replicating systems proved 
to be a fruitful ground for research for many years. A detailed review of this field with many additional references is contained in \cite{Owens}, 
\cite{M-K3}, and  \cite{F-M}. Pierre Kabamba introduced generation theory in \cite{kabamba} to analyze self-replicating systems. 

In this paper, which originates from \cite{Gi1}, I develop a new, much simpler, version of generation theory, which still captures all the details necessary for studying self-replication. 

Once a robot colony which is capable of self-replicating is designed, a further effort must be made to identify its optimal seeds.  A seed is a subset of the colony capable of manufacturing the whole colony. The optimization parameters might include the number of robots, the mass of robots, the cost of building robots on an extraterrestrial planet versus shipping them there, the time needed to manufacture robots at the destination, etc.  Menezes and Kabamba gave a background and several algorithms for determining optimal seeds for some special cases of such colonies in \cite{M-K1}, \cite{M-K2}, and \cite{M-K3}. They also gave a lengthy discussion of the merits of the problem and an extensive reference list in \cite{M-K3}. 

In this paper, the new version of the generation theory, developed by the author, is used to arrive at a remarkably simple new solution to the optimal seeding problem.

\section{A New Definition of a Generation System}

In the framework of generation theory, the entities that can potentially reproduce are called \textit{machines}, regardless of their physical nature (e.g., robots, microbes, or lines of computer code). Reproduction is achieved by the action of  machines on available resources, producing an outcome that may or may not be a machine itself.

Menezes and  Kabamba in \cite{M-K3} define a generation system as a quadruple $$(U,M,R,G)$$ where 
\begin{enumerate}
\item $U$ is a universal set that contains machines, resources, and outcomes of attempts at reproduction,
\item $M\subseteq U$ is a set of machines.
\item $R\subseteq U$ is a set of resources that can be used for reproduction, each resource is an ordered list.
\item $G:M \times R\rightarrow U$ is a generation function that maps a machine and
a resource ordered list into an element in the universal set.
\end{enumerate}

In this paper I introduce the following alternative definition of a generation system.

Note that an outcome of generation might not be a machine, however, I am interested only in outcomes which are machines, so I do not need to introduce the set $U$. If the outcome of generation is not a machine, I can define it to be the empty set.

Hence, I define a generation system to be a triple $$(M, R, G)$$ 

I do not require a resource to be an ordered list. Menezes and Kabamba needed resources to be ordered lists in order to run their algorithms in 
\cite{M-K3}.

I consider a more general generation function, which maps a subset of the set of machines and a subset of the set of resources into an outcome. I consider only outcomes which are subsets of the set of machines. An outcome might be the empty set.

This generalization of generation function is motivated by the observation that several robots might work simultaneously in order to produce
a new robot, and a family of robots, working together, might produce more than one robot. Hence I define a generation function to be a function of the following form:

$$G: 2^M \times 2^R \rightarrow 2^M$$ where $2^S$ denotes the set of all subsets of the set $S$.

I allow cannibalization, i.e. certain machines after fulfilling their function can be taken apart for parts or materials, so they become resources. Hence it is possible that $M \cap R \neq \emptyset $. However, I do not allow the empty generation situation when a resource becomes a machine without any transformation. Hence if a machine $m_0$ is generated by a set of machines $M_0 \subset M$ acting on a set of resources $R_0 \subset R$, I require that $m_0 \notin R_0$. Menezes and Kabamba in \cite{M-K3} call such generation systems \textit{weakly regular}.

A set of machines $M_0 \subseteq M$ is called self-replicating if there exists a set of resources $R_0 \subseteq R$ such that $G(M_0,R_0) = M_0$.

\section{Formulation of the Optimal Seeding Problem}

The general problem of seeding of a space colony can be put in the following simple form, where it becomes a transportation problem.

A set $M$ of machines must be delivered to a certain destination. I assume that the set $M$ is self-replicating. The set $M$  exists at the base, so one solution of the problem is to load  $M$ at the base on some carrier and to transport it to the destination. However, transportation is expensive, and my  budget is limited, so I must transport as little as possible. Note that  for  a mining colony, all of the resources $R$ for self-replicating of $M$ must be available at the destination except, possibly, for some machines in $M$ which become resources after fulfilling their tasks. 

For all practical purposes I can assume that both sets $M$ and $R$ are finite.

I  assume that for any subsets $M '\subseteq M$ and $R' \subseteq R$  the generation function $G(M',R')= M'' \subseteq M$ is given. I need the generation function $G(M',R')$ to be defined for all subsets $M'\subseteq M$ and $R' \subseteq R$ in order to run the algorithm described in the next section of the paper. In addition, I assume that a cost function $T(G(M',R'))$, which is vector-valued is given. The first component of $T(G(M',R'))$ provides the cost of running the generation function on the set $(M',R')$, the remaining components of the function $T(G(M',R'))$ might be monetary cost of the manufacturing process, time required for the process, the energy consumption, etc. 

I also assume that for any subset $M '\subseteq M$ a cost function $U(M')$, which also might be vector-valued, is given. The first component of $U(M')$ provides the cost of transporting all the machines in $M'$ to the destination.

I want to find all seeds of $M$, i.e. subsets  of $M$ capable of generating $M$. Seeds exist, because I assumed that $M$ is self-replicating, so $M$ is a seed of itself. In general, $M$ might be the only seed of itself, however, I hope to find a number of different  seeds.

Moreover, I want to find a seed optimal with respect to $T(G)$ and to $U$  (or to a some subset of components of $T(G)$ and $U$), and transport it to the destination. If there are several optimal seeds, I can choose the optimal seed randomly out of a set of several candidates.

\section{A New Solution to the Optimal Seeding Problem}

I assume that the sets $M$ and  $R$ are finite. 

I construct the following directed labeled graph $\Gamma$. 

The set of vertices of $\Gamma$ is $2^M$. Two vertices $M_1$ and $M_2$ are connected by a directed edge if there exists a set of resources
$R_1 \subseteq R$ such that $M_2 =G(M_1, R_1)$ and $R_1 \cap M_2 =\emptyset$. Note that there might exist several distinct sets $R_1$, satisfying the 
above description. However as the set $R$ is finite, there might be only finitely many such $R_1$. Denote these sets by 
$R_{1,1}, R_{1,2}, \cdots , R_{1,n}$. Hence any pair of vertices $M_1$ and $M_2$ in $\Gamma$ might be connected by finitely many directed edges. 
An edge of $\Gamma$ connecting vertex $M_1$ to $M_2$ is labeled by the function $T(G(M_1, R_{1,i}))$. 

As $\Gamma$ is a directed graph, a directed path $p$ in $\Gamma$  is a finite sequence of directed edges $e_1 \cdots e_{n_p}$ such that the 
terminal vertex of any edge $e_j$ (except for the last one) coincides with the initial vertex of the next edge $e_{j+1}$. 
For any vertex $M'$ in $\Gamma$, I  define its strongly connected component $\Gamma_{M'}$ to be the set of vertices in $\Gamma$ which can be connected to $M'$ by a  directed path. By construction, $\Gamma_M'$ is the set of all seeds of the set $M'$. In particularly, the strongly connected component of the vertex $M$ is the set of all seeds of $M$.

In order to find a seed of $M$ with minimal transportation cost, I consider the set of all simple (without self-intersections) directed  paths, (using any standard algorithm for the purpose \cite{A-H-U}), connecting a vertex in $\Gamma_M$ to $M$. For any such path $P= e_1,  e_2, \cdots e_{n_p} $ let $M'_i$ be the initial vertex of the edge $e_i$ and let $T(G(M'_i, R_i))$ be the label of the edge $e_i$. Let  $T_1(G(M'_i, R_i))$ be the first component of the function  $T(G(M'_i, R_i))$, which provides the cost of runing the generation function  $G(M'_i, R_i)$, and let $U_1(M'_1)$ be the first component of the function $U(M'_1)$, which provides the cost of transporting  all the machines in $M'_1$ to the destination, (note that $M'_1$ is the initial vertex of the path $P$). 

The cumulative cost function for the path $P$ is given by the following sum
$$Cost(P)= U_1(M'_1) + T_1(G(M'_1, R_1)) + T_1(G(M'_2, R_2)) + \cdots + T_1(G(M'_{n_p}, R_{n_p}))$$

As the set $\Gamma_M$ is finite, I  can create a list of vertices of $\Gamma_M$ sorted by the value of the function $Cost(P)$  and determine the seeds with minimal transportation cost.

\section{Discussion}

This paper describes a novel approach to the problem of identifying an optimal seed of a self-
reproducing automaton. This approach is based on generation theory, a mathematical construct
capable of describing arbitrary matter and information processing systems in sufficient detail for
studying the capacity of such systems for self-reproduction. This paper introduces a much
simpler version of generation theory that still captures all the details necessary for studying self-
reproduction. This new theory is  used to arrive at a remarkably simple solution to the
optimal seeding problem.

The main contribution of this paper is the novel abstraction it introduces, namely the
simplified version of the generation theory. This contribution is significant. The provided solution
to the optimal seed problem reinforces the significance of the result and illustrates its practical
use. 

\section{Conclusions}

The new solution to the optimal seeding problem, introduced in this paper, demonstrates that the new definition of the generating system, given in this paper, is a powerful mathematical tool. In a  forthcoming paper \cite{Gi2}, I shall describe the mathematical parallels between the new version of generation theory and certain well-studied objects of classical mathematics.

\section{Acknowledgment}

I want to thank Pierre Kabamba for introducing me to the initial version of the generation theory and Galip Ulsoy for helpful discussions.

\end{document}